\documentclass[runningheads]{llncs}
\usepackage[T1]{fontenc}
\usepackage{graphicx}
\usepackage{amsmath}
\usepackage{hyperref}
\begin{document}
\title{Survey of Large Multimodal Model Datasets, Application Categories and Taxonomy}
%
%
\author{Anonymous PAKDD 2025 Submission}
\author{Priyaranjan Pattnayak\inst{1}\and
Hitesh Laxmichand Patel\inst{2} \and
Bhargava Kumar\inst{3} \and
Amit Agarwal\inst{4} \and
Ishan Banerjee\inst{5} \and
Srikant Panda\inst{6}
Tejaswini Kumar\inst{7}}

\authorrunning{P. Pattnayak et al.}

\institute{University of Washington, Seattle, USA \and
New York University, New York, USA \and
Columbia University, New York, USA\and
Liverpool John Moores University, Liverpool, UK \and
Chennai Mathematical Institute, Chennai, India \and
Birla Institute of Technology, Pilani, USA \and
Columbia University, New York, USA\\
}

\maketitle              
\begin{abstract}
Multimodal learning, a rapidly evolving field in artificial intelligence, seeks to construct more versatile and robust systems by integrating and analyzing diverse types of data, including text, images, audio, and video. Inspired by the human ability to assimilate information through many senses, this method enables applications such as text-to-video conversion, visual question answering, and image captioning. Recent developments in datasets that support multimodal language models (MLLMs) are highlighted in this overview. Large-scale multimodal datasets are essential because they allow for thorough testing and training of these models. With an emphasis on their contributions to the discipline, the study examines a variety of datasets, including those for training, domain-specific tasks, and real-world applications. It also emphasizes how crucial benchmark datasets are for assessing models' performance in a range of scenarios, scalability, and applicability. Since multimodal learning is always changing, overcoming these obstacles will help AI research and applications reach new heights.

\end{abstract}
\section{Introduction to Multimodal Learning and Large Language Models}

Multimodal learning, a growing field in AI, focuses on integrating and processing multiple data types like text, images, and audio, aiming to replicate human cognition, which naturally combines sensory inputs. This approach enables more robust and intelligent systems compared to single-modality methods.

Large language models (LLMs) such as GPT-3, BERT, and T5 excel in text-based tasks like question answering and summarization \cite{ref1}. However, they struggle with non-text data, driving interest in multimodal large language models (MLLMs) that combine LLMs’ language capabilities with computer vision's strengths. MLLMs have achieved state-of-the-art results in tasks like image captioning and visual question answering \cite{ref2}. Challenges remain, including limited high-quality datasets, high computational costs, and ethical concerns like bias and privacy \cite{ref11}. Despite these hurdles, MLLMs hold transformative potential in healthcare, education, and research, making them a key focus in advancing AI.

\subsection{Multimodal Learning: Foundations and Concepts}

\begin{figure*}[h!]
		\centering
		\includegraphics[width=\textwidth]{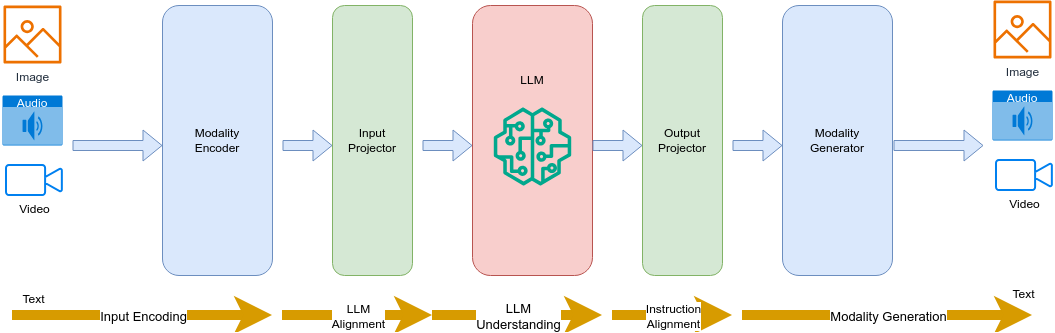} 
		\caption{Flowchart describing the multimodal language model pipeline }
		\label{fig:multimodal_learning2}
        \vspace{-1em}
	\end{figure*}

Multimodal learning consists of building models that can process and combine information from various data modalities, such as text, images, audio, and video. This is due to the fact that real-world experiences are inherently multimodal in nature, and the different types of information carried by the various modalities provide a way to understand such complex environments more thoroughly \cite{ref11}.

Multimodal learning incorporates multiple types of data: texts, images, audio, and video. These create representations, each unique to one modality. Given the diverse nature of various types of data, different methods have traditionally been used to capture their properties. Text, for example, is represented with word embeddings emphasizing meaning and structure \cite{ref12}, while image data most often depend on a convolutional neural network that would extract details from visual scenes. Similarly, audio data is often transformed into spectrograms or mel-frequency cepstral coefficients to capture patterns in time and frequency \cite{ref13}. A typical pipeline for a large multimodal model (MLLM) is shown in Fig. \ref{fig:multimodal_learning2}, where inputs are first processed through a modality encoder to unify their representations. These are then refined by an input projector and passed into a Large Language Model (LLM) for deeper alignment and understanding. Finally, the output projector and modality generator transform the model's results into meaningful outputs, enabling tasks like generating multimodal content or translating between data types.

Fusion of modality representations is a key focus in multimodal learning. The widely used methods are early fusion, in which the concatenation or combination of representations is carried out at an initial stage of processing \cite{ref14}, and late fusion, where modality-specific representations are combined later in the process, often with attention or gating mechanism \cite{ref14}.

Beyond representation and fusion, there are more challenges involved with multimodal learning, like alignment, translation, and co-learning. Alignment allows temporal or semantic synchronization across modalities, which is an essential task for video understanding or audio-visual speech recognition \cite{ref16}. Translation helps in modality transformation, as generating images from text, for example \cite{ref17}. Co-learning allows to learn in the conditions when some of the data modalities are not available or damaged, by transferring knowledge from available modalities \cite{ref13}.

Recent progress on LLMs, such as BERT, GPT, and DALL-E, has considerably accelerated progress in multimodal learning. These models are very good at both understanding and generating text; their extension to multiple data types now also enables answering questions about images, creating picture descriptions, or even generating images based on text \cite{ref19}.

In short, multimodal learning has been a very critical factor in the development of intelligent systems that effectively process and integrate information from varied sources. The complementary strengths of multiple modalities ensure that this area continuously creates innovation across domains like NLP, computer vision, and robotics, among others, with ever-widening scopes of applications and research directions.

\subsection{Multimodal Large Language Models: Opportunities and Challenges}

Recent advances in LLMs have laid the path for multimodal large language models that combine data across modalities, such as text, images, audio, and video \cite{ref3}. MLLMs hold the potential to transform various domains by enhancing understanding and representation through a mix of different modalities.

MLLMs expand the capability of LLMs to wider ranges of tasks beyond traditional text-only models. This class of models is very strong on tasks like image captioning, visual question answering, and text-to-video generation-all requiring an in-depth understanding of language-visual relationships \cite{ref5}.

The integration of multi-modal data opens the way for scientific research and domain-specific applications for MLLMs by pushing the boundary. Some critical domains like medical imaging, autonomous driving, geospatial intelligence combine textual, visual, and sensor data to yield more realistic decision-making processes.

Despite the potential of MLLMs, there are significant challenges in developing them. Among the primary issues is the absence of large-scale, high-quality multimodal datasets \cite{ref30}. Complex, unbiased data covering the richness of reality is a necessary ingredient to train robust MLLMs\cite{ref11}.

Another challenge is the increase in computational demands and complexity in integrating these various modalities. The training and deployment of MLLMs require considerable resources; thus, there is a need to develop novel model architectures, efficient training strategies, and hardware capabilities \cite{ref11}.

Finally, ensuring the reliability, interpretability, and ethical alignment of MLLMs is important. With increased sophistication of these models, there is a growing need to provide insights into their decision-making processes to reduce biases and align them more closely with human values. The development of robust evaluation frameworks and interpretability tools are necessary to engender trust in MLLMs \cite{ref32}.

Despite this, the prospects for MLLMs are enormous. While incorporating multimodal data, the models pave the way for a better comprehension of complicated scenarios, hence giving birth to new applications and promoting scientific research accordingly. In addition, future interdisciplinary collaboration and emphasis on ethical considerations are the critical elements toward the transformation that can be brought by MLLMs \cite{ref11}.

\begin{figure}[h!]
    \centering
    \includegraphics[width=\textwidth]{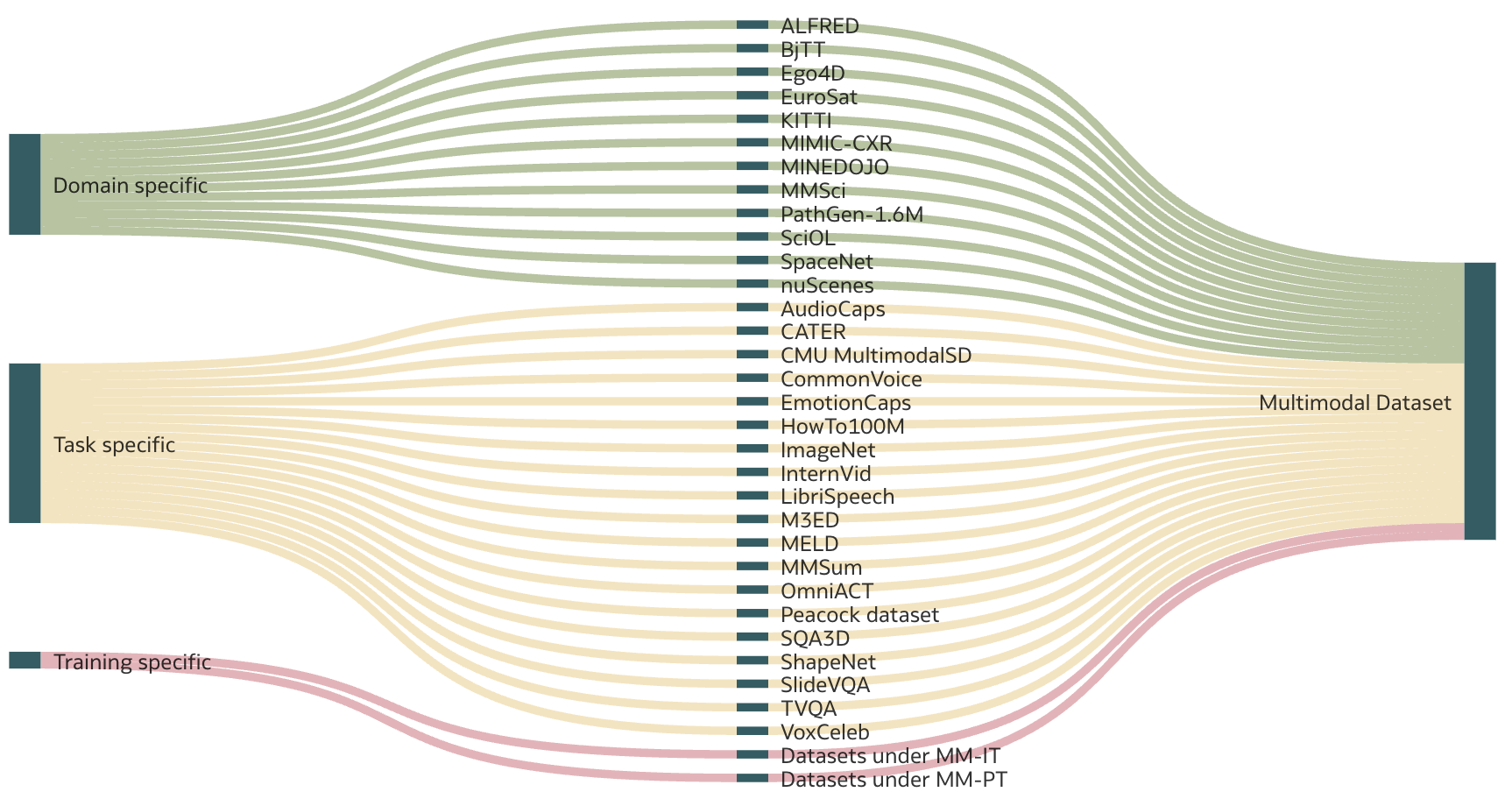} 
    \caption{An illustration representing the high-level classification of the datasets mentioned in the survey under Training specific (datasets under MM-IT and MM-PT), Task specific and Domain specific.}
    \label{fig:flowchart}
    \vspace{-1em}
\end{figure}

In the following sections, we classify datasets that are critical for MLLMs into three major types: training-specific datasets, task-specific datasets, and domain-specific datasets, as illustrated in Fig. \ref{fig:flowchart}.

\section{Multimodal Datasets for Training Specific Needs}
The development of multimodal datasets is essential to advancing MLLMs. These datasets span a wide range of modalities and applications, enabling researchers to train models that can integrate and reason across various data types. Challenges in dataset design include ensuring scale, modality diversity, annotation quality, and applicability to real-world scenarios. In this section, we discuss few datasets that are used for training (both pre-training and instruction tuning) multimodal models, as illustrated in Fig. \ref{fig:flowchart} under training specific types.

\begin{figure*}[t!]
     	\centering
     	\includegraphics[width=0.8\textwidth]{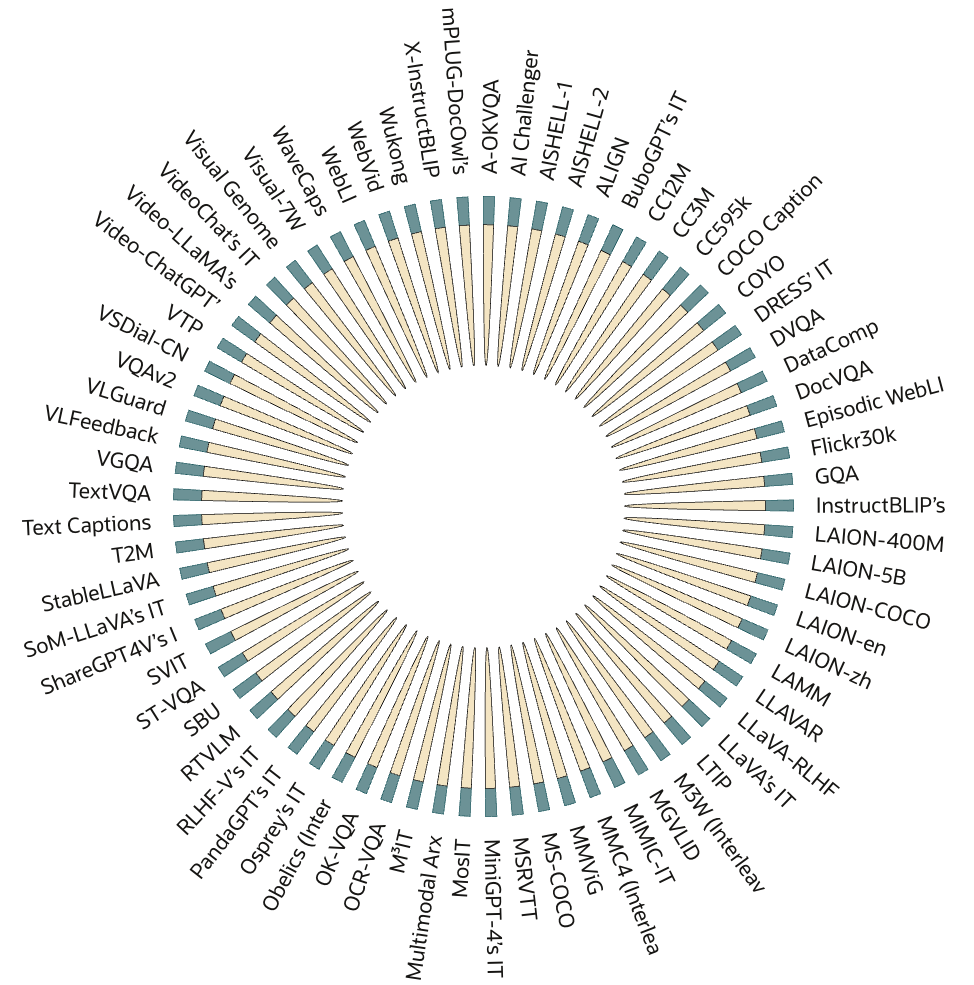} 
     	\caption{The datasets released as part of MLLMs}
     	\label{fig:fig_mllm}
        \vspace{-1em}
     \end{figure*}

To train large-scale multimodal models for tasks like picture captioning, visual question answering, and audio-text understanding, the MLLMs dataset \cite{ref3} consists of text, images, video, and audio. For academics looking into multimodal integration, its scope across several modalities offers a cohesive framework. In Fig. \ref{fig:fig_mllm}, datasets made available as part of MLLM models are displayed. 

Multimodal Pre-Training (MM-PT) and Multimodal Instruction Tuning (MM-IT) are the two primary tenets of the MLLM development pipeline. Image-Text, Video-Text, and Audio-Text datasets are utilized in these phases and are further separated into:
\begin{itemize}
    \item Image-Text Pairs: Simple $<$img1$>$ $<$txt1$>$ format.
    \item Interleaved Image-Text Corpus: Mixed sequences like $<$txt1$>$ $<$img1$>$ $<$txt2$>$ $<$txt3$>$ $<$img2$>$.
\end{itemize}

The process of pre-training a model to comprehend and align modalities, including text and a picture, is known as MM-PT. Along with offering a basis for comprehending and aligning various data kinds, MM-PT enables the capture of links between various modalities, such as connecting verbal descriptions to visual attributes. This is crucial for activities like visual reasoning or image captioning. MM-IT is an approach that uses datasets formatted as instructions to adapt pre-trained MM-LLMs. In exploring the landscape of datasets, we leveraged information from existing literature \cite{ref139,ref133,ref132}. Table \ref{tab:tbl_mmpt} provides a comprehensive list of such datasets used for MM-PT, including modality, size, and year of release. 

    	\begin{table*}[htb]
    	\centering
        \caption{Overview of datasets under MM-PT.}
    \label{tab:tbl_mmpt}
    	\resizebox{1.0\linewidth}{!}{
    		\begin{tabular}{l c c c c c}
    			\hline
    			\textbf{Dataset} & \textbf{Modality} & \textbf{Description} & \textbf{Size} & \textbf{Year}  \\ \hline
    			ALIGN & Image & Large-scale image-text pairs for training vision-language models & 1.8B pairs & 2021\\
    			LTIP & Image & Large-scale image-text pairs dataset & 312M pairs & 2022\\
    			Visual Genome & Image & Dataset with object relationships and attributes for VQA & 108K images & 2017\\
    			CC3M & Image & Conceptual Captions dataset & 3.3M pairs & 2018\\
    			CC12M & Image & Large-scale captioning dataset for vision-language tasks & 12.4M pairs & 2021\\
    			SBU & Image & Image-caption pairs dataset & 1M pairs & 2011\\
    			LAION-5B & Image & Large-scale web-crawled image-text pairs & 5.9B pairs & 2022\\
    			LAION-400M & Image & Subset of LAION dataset with 400M image-text pairs & 400M pairs & 2021\\
                MS-COCO & Image & Object detection, segmentation, and captioning dataset & 124K images & 2014\\
    			YFC100M  & Image & Large-scale collection of web-crawled images and videos & 99.2M pairs & 2014\\
    			LAION-COCO & Image & COCO-aligned subset of LAION dataset & 600M pairs & 2022\\
    			Flickr30k & Image & Image-captioning dataset with real-world scenes & 31K images & 2014\\
    			COYO-700M & Image & Large-scale web-crawled dataset for image-text pairs & 747M pairs & 2022\\
    			Wukong & Image & Chinese language image-text pairs dataset & 101M pairs & 2022\\
    			WebLI & Image & Large-scale dataset for multilingual image-text pairs & 10B images & 2022\\
                    FILIP   & Image & Fine-grained vision-language dataset & 300M images & 2021\\
                    CLIP   & Image & Large-scale image-text pairs, collected from diverse sources & 400M images & 2021\\		
    			WebVid & Video & Large-scale video dataset with captions & 10M videos & 2021\\
                    Youku-mPLUG  & Video & Large-scale Chinese Video-Language Dataset & 10M videos & 2023\\
                    VideoCC3M & Video & Large-scale Video-Language Dataset & 6.3M videos & 2022\\
                    YouTube8M & Video & Large and Diverse Labeled Video Dataset for Video & 8M videos & 2022\\
                    Vript & Video & Large and fine-grained video-text dataset  & 420K videos & 2024\\
                    OpenVid-1M  & Video & Large-Scale High-Quality Dataset for Text-video  & 1M videos & 2024\\
                    VidGen-1M  & Video & Large-Scale High-Quality Dataset for Text-video  & 1M videos & 2024\\
    			VTP & Video & Dataset with video-text pairs & 27M videos & 2022\\
                    AudioSet  & Audio & Dataset with audio-video pairs & 2M audio-videos & 2011\\
                    YODAS  & Audio & Dataset with audio-text pairs & 500 K audio & 2024\\
                    LAION-Audio  & Audio & large-scale audio-text dataset  & 630K audio-text & 2023\\
                    AVSET-10M & Audio & large-scale audio visual dataset  & 10M audio-video & 2023\\
                    Auto-ACD  & Audio & large-scale audio-text dataset  & 1.5M audio-text & 2023\\
    			AISHELL-1 & Audio & Chinese language audio dataset for ASR & 128K utterances & 2023\\
    			AISHELL-2 & Audio & Expanded AISHELL audio dataset for ASR & 1M utterances & 2023\\
    			WaveCaps & Audio & Dataset with audio-caption pairs & 403K pairs & 2023\\
    			VSDial-CN & Image, Audio & Chinese language visual dialogue dataset & 1.2M audio utterances & 2024\\
    			
    	\end{tabular}}
    
    \end{table*}

The instruction-formatted datasets used for tuning in the MM-IT stage improve zero-shot performance by increasing task flexibility using the following non-exhaustive list of methods:

\begin{itemize} 
\item Supervised Fine Tuning (SFT) improves MM-LLMs on visual question answering through the conversion of data into templates that may be utilized-for example, single-turn or multi-turn.
\item Reinforcement Learning with Human Feedback (RLHF) aligns model responses with human purpose by using natural language input.
\end{itemize}

\subsubsection{Notable Datasets in MM-PT and MM-IT}
\begin{itemize}
    \item 
\textbf{LAION-5B:} The LAION-5B\cite{ref115} dataset contains 5.85 billion CLIP-filtered image-text pairs (2.32 billion in English) supporting large-scale multi-modal research with tools for watermark detection and NSFW filtering.
    \item 
\textbf{MS-COCO:} The MS-COCO\cite{ref110} contains over 330K images with five human-written captions each, serving as a benchmark for image recognition, segmentation, and captioning.
    \item 
\textbf{Flickr30k:} Flickr30k\cite{ref105} contains 31,000 images, each annotated with five captions, for image captioning and cross-modal retrieval.
    \item 
\textbf{COYO-700M:} The COYO-700M\cite{ref139} dataset includes 747 million image-text pairs, extracted from Common Crawl and filtered for quality for model training.
    \item 
\textbf{CC12M:} CC12M(Conceptual Captions 12M)\cite{ref113} is a large dataset with 12M image-text pairs for training vision-language models. It uses web images and their captions, offering varied data to support robust  learning.
    \item 
\textbf{OpenVid-1M:} OpenVid-1M\cite{ref126} is a dataset of 1M high-quality text-video pairs, created to support text-to-video generation research.
    \item 
\textbf{AISHELL-2:} AISHELL-2 \cite{ref137} is a Mandarin speech corpus with 1 million transcribed segments from 400 speakers, offering high-quality data for automatic speech recognition tasks.
    \item 
\textbf{LAMM:} It provides 186K language-image and 10K language-3D instruction-response pairs for MLLMs, with a modular framework for 2D and 3D vision tasks \cite{ref37}.
\item 
\textbf{Wukong:} With 100 million image-text pairs, the Wukong dataset is a comprehensive Chinese multimodal resource that focuses on vision-language pretraining \cite{ref133}.
\item 
\textbf{VRIPT:} Each of the 420,000 clips and 12,000 high-resolution videos in the VRIPT dataset \cite{ref125} has a comprehensive caption that describes the content, shot kinds, and camera motions. It is intended for sophisticated video comprehension and production, including transcriptions.
\item
\textbf{SBU:}The SBU Caption Dataset consists of 1 million image-text pairs, where image captions are sourced from user-provided descriptions. It is widely used for training and evaluating vision-and-language models\cite{ref132}.
    \item 
\textbf{LLaVA-Instruct:} The dataset contains 150K set of GPT-generated multimodal instruction-following data to train LLaVa\cite{ref136} model. 
    \item 
\textbf{SVIT:} SVIT\cite{ref135} contains 4.2 million examples for visual captioning and QA tasks, created with GPT-4 and manual annotations, enabling models like SVIT-v1.5 to excel in visual understanding benchmarks.
    \item 
\textbf{MIMIC-IT:} MIMIC-IT\cite{ref134} includes 2.8 million multimodal samples for instruction-following tasks across multiple languages, supporting models like OpenFlamingo in understanding complex multimodal instructions.
    \item 
\textbf{SoM-LLaVA's IT:} SoM-LLaVA's\cite{ref138} Instruct dataset consists of 695,000 image-text pairs with structured visual prompts, enhancing model reasoning in tasks like visual question answering using set-of-mark prompting.

\end{itemize}

We utilized information from existing literature \cite{ref139,ref132} to curate the list of MM-IT datasets. Table \ref{tab:tbl_mmit} shows a comprehensive list of all such datasets that have been used for MM-IT, including modality, size, and year of release. These further align with human purpose by feedback and instruction-aware formatting to develop better interaction capabilities for the MLLMs. 

    \begin{table*}[h]
    	\centering
        \caption{Overview of datasets under MM-IT.}
    	\label{tab:tbl_mmit}
    	\resizebox{\textwidth}{!}{
    		\begin{tabular}{l l c l r}
    			\hline
    			\textbf{Dataset} & \textbf{Modality} & \textbf{Description} & \textbf{Size} & \textbf{Year}\\
    			\hline
                 MiniGPT-4 IT & Image + Text & Dataset for training generative models with images and text pairs. & 5K & 2023\\
                StableLLaVA & Image + Text & Robust dataset for vision-language tasks with diverse image-text pairs. & 126K & 2023\\
                Q-Pathway & Image + Text &  Dataset for improving low-level visual perception. & 19K & 2023\\  
                LLaVA-Instruct-150K & Image + Text & Image-text instruction tuning dataset. & 150K & 2023\\
                ShareGPT4V's IT & Image + Text & Interactive dataset integrating multiple visual sources with text. & 100K & 2023\\
                MultiModalGPT IT & Image + Text & Dataset designed to visual inputs. & 6K & 2023\\
                SoM-LLaVA's IT & Image + Text & Dataset for fine-tuning models on complex visual prompts. & 695K & 2024\\
                VideoChat's IT & Video + Text & Dataset for training conversational agents with video inputs. & 11K & 2023\\
                Video-ChatGPT's IT & Video + Text & Dataset for generating dialogues based on video content. & 100K & 2023\\
                Video-LLaMA's IT & Image/Video + Text & Integrative dataset combining image and video for text generation. & 171K & 2023\\
                InstructBLIP's IT & Image/Video + Text & Comprehensive dataset for instruction-based visual understanding. & 1.6M & 2023\\
                X-InstructBLIP's IT & Image/Video/Audio/3D + Text & Dataset combining multiple modalities for complex interactions. & 1.8M & 2023\\
                MIMIC-IT & Image/Video + Text & Medical dataset integrating image and video data with annotations. & 2.8M & 2023\\
                VSTaR-1M & Image/Video + Text & Video instruction tuning dataset. & 1M & 2023\\
                PandaGPT's IT & Image + Text & Dataset for training generative conversational models with visuals. & 160K & 2023\\
                MGVLID & Image + Bounding Box + Text & Visual dataset with bounding box annotations for training. & 108K & 2023\\
                M³IT & Image/Video/Bounding Box + Text & Multi-modal dataset for rich visual-text interactions. & 2.4M & 2023\\
                LAMM & Image + 3D + Text & Dataset focusing on 3D interactions and image-text pairs. & 196K & 2023\\
                BuboGPT's IT & (Image + Audio)/Audio + Text & Mixed dataset for training with audio and visual inputs. & 9K & 2023\\
                mPLUG-DocOwl'sIT & Image/Table/Web + Text & Dataset for document understanding across various formats. & -- & 2023\\
                T2M & Text to Image/Video/Audio + Text & Transformative dataset for generating visuals from text prompts. & 14.7K & 2023\\
                MosIT & Image + Video + Audio + Text & Complex dataset for multi-modal training and interactions. & 5K & 2023\\
                Osprey's IT & Image + Text & Diverse dataset for fine-tuning models with image inputs. & 724K & 2023\\
                LLaVA-RLHF & Image + Text & Dataset focusing on reinforcement learning from human feedback. & 10K & 2023\\
                RLHF-V's IT & Image + Text & Dataset designed for video input interactions in RL training. & 1.4K & 2023\\
                Shikra & Image + Text & Language vision question answeing dataset & 156K & 2022\\
                Vision-Flan & Image + Text & Language vision instruction tuning dataset & 1.6M & 2024\\
                RTVLM & Image + Text & Recent dataset focusing on real-time visual language modeling. & 5K & 2024\\
                LVIS-Instruct4V & Image + Text & fine-grained visual instruction dataset. & 220K & 2024\\
                SVIT & Image + Text & Large scale visual instruction dataset. & 4.2M & 2024\\
                MultiInstruct & Image + Text & Multi-modal dataset for instruction tuning. & 235K & 2023\\
                MAmmoTH-VL-Instruct & Image + Text & Multi-modal dataset for instruction tuning. & 12M & 2024\\               
                Leopard-Instruct & Image + Text & text rich dataset for instruction tuning. & 925K & 2023\\
                MMInstruct & Image + Text & High-Quality data Extensive Diversity & 973K & 2024\\
    		
    		\end{tabular}
    	}
    	
    \end{table*}

\section{Multimodal Datasets for Task Specific Applications}

\begin{figure*}[t!]
    \centering
    \includegraphics[width=0.7\textwidth]{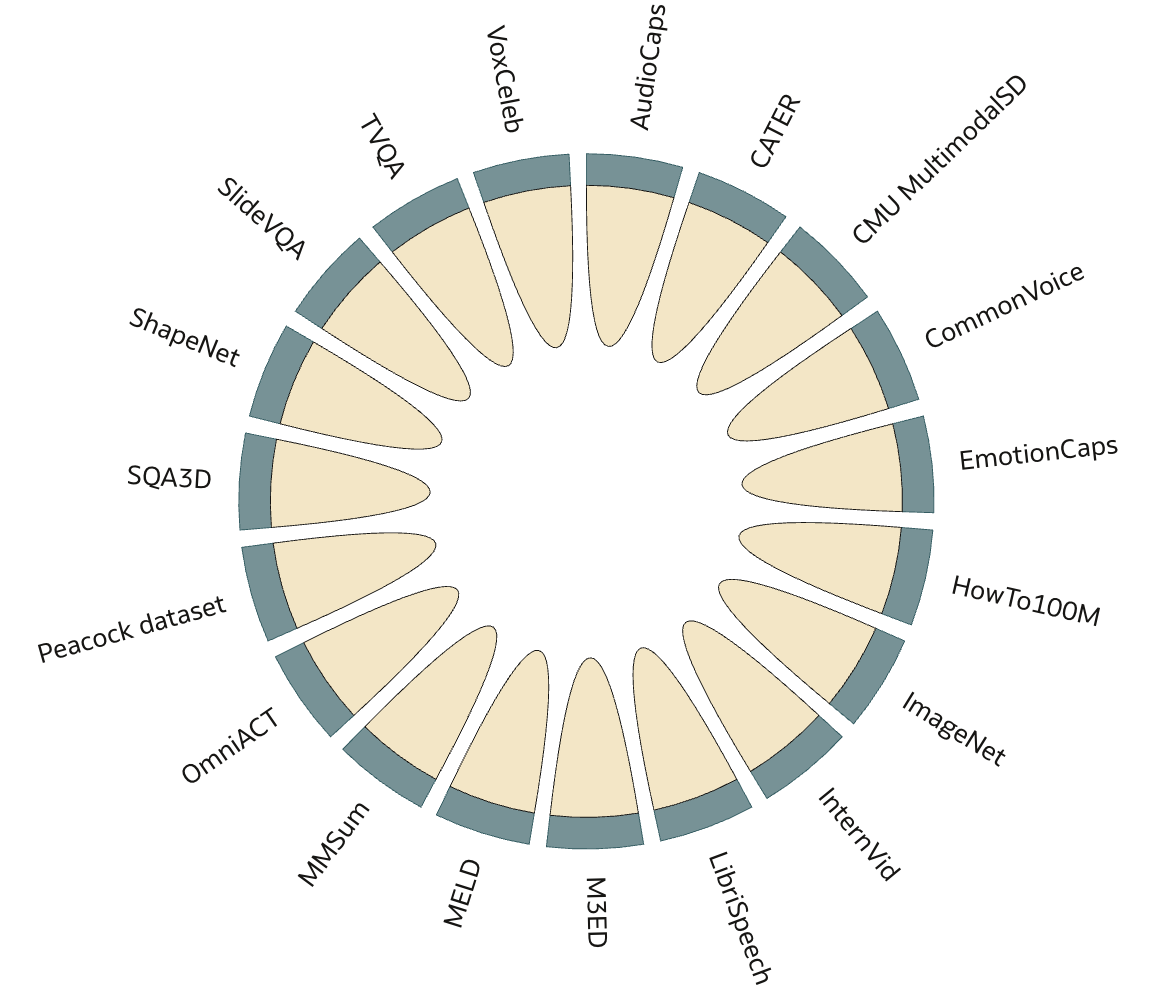} 
    \caption{An illustration of the datasets as per the survey under Task Specific Needs}
    \label{fig:fig_task}
    \vspace{-1em}
\end{figure*}

The datasets discussed in this section form the core of designing versatile models that can work on a wide range of tasks, including sentiment analysis, emotion detection, and visual question answering. They form the key building blocks for robust and adaptable multimodal frameworks. Several of these datasets are highlighted in Fig. \ref{fig:fig_task}, while a detailed overview is given in Table \ref{tab:tbl_task_specific}, including dataset modality, size, and release year.

\subsubsection{SlideVQA:}

SlideVQA \cite{ref92} expands on single-image VQA datasets by introducing a multi-image framework for complex reasoning tasks. It includes over 2,600 slide decks (52,000+ images) and 14,500+ questions, focusing on multi-hop reasoning and numerical analysis with annotated arithmetic expressions. The dataset presents challenges in real-world document understanding, such as evidence identification and sequence-based reasoning across multiple images.

\subsubsection{Peacock dataset:}

The Peacock dataset \cite{ref7} addresses a gap in multilingual, culturally relevant multimodal resources by focusing on the Arabic language and its cultural context. It enables the training of multilingual large language models (MLLMs) tailored for Arabic-speaking populations, combining text, visuals, and culturally significant elements. This enhances the models' applicability across various languages and cultures.

\subsubsection{OmniACT:}

OmniACT \cite{ref93} serves as a benchmark for evaluating agents' ability to generate programs and test models for computer task automation. It includes a wide range of tasks, from simple commands like "Play the next song" to complex ones like "Send an email to John Doe with meeting details." The dataset combines screen visuals with task descriptions, supporting studies in automated task execution through visually grounded interfaces.

\subsubsection{InternVid:}

InternVid \cite{ref90} is a large-scale video-text dataset with 7 million videos (760,000 hours, 234 million snippets, and 4.1 billion words) designed for learning video-text representations. Leveraging the ViCLIP model and contrastive learning, it excels in zero-shot action recognition, video retrieval, and text-to-video/video-to-text tasks. This dataset significantly advances video-based multimodal tasks, including dialogue systems.

\subsubsection{ImageNet:}
Tens of millions of annotated photos (150Gb) that are in line with the WordNet semantic hierarchy are available through ImageNet \cite{ref95}, a large-scale image ontology. Applications that show the value of sizable, well-organized image datasets include object identification, image categorization, and clustering.

\subsubsection{CMU MultimodalSD dataset:}

The CMU MultimodalSD dataset \cite{ref71} contains over 1,000 YouTube video clips, each accompanied by text transcripts and sentiment annotations. This dataset is designed for multimodal sentiment analysis, where visual and textual information are integrated to train models for predicting sentiment in real-time video scenarios.

\subsubsection{MELD:}

The MELD dataset \cite{ref39} contains 13,000+ utterances from 1,433 dialogues, annotated into seven emotion categories. It integrates acoustic and textual data for emotion recognition in real-world conversations. The dataset ensures accuracy through timestamp alignment and majority-vote annotation. Compared to EmotionLines, it reduces dialogues by constraining timestamps and shared scenes. After resolving annotation disagreements, final dataset consists of audiovisual clips, 16-bit PCM WAV audio, and aligned visual, audio, and text modalities.

\subsubsection{HowTo100M:}
The HowTo100M dataset \cite{ref80} contains 136 million video clips sourced from over 1.22 million narrated instructional videos, spanning more than 23,000 tasks. Models trained on this dataset excel in text-to-video retrieval and action localization, outperforming benchmarks like YouCook2 and CrossTask, and generalize effectively to other datasets, including MSR-VTT and LSMDC, after fine-tuning. The dataset, along with associated tools and models, is openly available for research.

\subsubsection{AudioCaps:}
Using the AudioSet dataset \cite{ref89}, Audio Captioning in the Wild \cite{ref81} offers a collection of 46K audio clips with crowdsourced human-generated captions. It investigates sophisticated captioning models and audio representation approaches. Notably, it adds two new elements to attention-based models for better captioning performance: aligned semantic attention and a top-down multi-scale encoder.

\subsubsection{CommonVoice:}
The Common Voice dataset \cite{ref82} is a large-scale, multilingual resource for Automatic Speech Recognition (ASR) and language identification, built through crowdsourcing. By November 2019, it included contributions from over 50,000 participants, amassing 2,500 hours of audio across 29 supported languages, with data collection underway for 38 languages. This makes Common Voice the largest publicly accessible audio corpus for speech recognition tasks.

\subsubsection{EmotionCaps:}
EmotionCaps \cite{ref83} introduces a dataset of 120,000 audio clips with synthetic descriptions enriched with emotional content, addressing the gap in emotion-aware audio-language pretraining. By incorporating emotional data, models trained on EmotionCaps generate captions that better reflect audio emotions, surpassing baseline models and driving research in audio captioning.

\subsubsection{M3ED:}
The M3ED dataset \cite{ref84} comprises 990 dialogues and 24,449 utterances from 56 TV series, annotated with seven emotions—happy, surprise, sad, disgust, anger, fear, and neutral—across acoustic, visual, and textual modalities. As the first Chinese multimodal emotional dialogue resource, it facilitates cross-cultural emotion analysis. Additionally, it introduces the MDI framework for context-aware emotion recognition, achieving strong performance with state-of-the-art methods.

\subsubsection{ShapeNet:}
ShapeNet \cite{ref85} is a large-scale repository of 3D CAD models, organized using the WordNet taxonomy and enriched with semantic annotations like part structures, symmetry planes, rigid alignments, and physical dimensions. Featuring over 3,000,000 indexed models, with 220,000 categorized into 3,135 WordNet synsets, it serves as a key resource for object visualization, geometric analysis, and benchmarking in computer graphics and vision research.

\subsubsection{SQA3D:}
Situated Question Answering in 3D Scenes (SQA3D) \cite{ref97} introduces a task to assess embodied agents' understanding of 3D environments through reasoning about spatial relations, commonsense, navigation, and multi-hop logic. Using 650 ScanNet scenes, the dataset includes 6.8k situations, 20.4k descriptions, and 33.4k questions. Results reveal a significant performance gap, with state-of-the-art models achieving 47.20\% accuracy compared to 90.06\% by humans, underscoring the challenge of 3D multi-modal reasoning.

\subsubsection{TVQA:}
TVQA \cite{ref41} is a multimodal dataset for video question answering, comprising 925 episodes from six TV shows across sitcoms (\textit{The Big Bang Theory, How I Met Your Mother, Friends}), medical dramas (\textit{Grey’s Anatomy, House}), and a crime drama (\textit{Castle}), totaling 461 hours. The dataset includes 21,793 clips, segmented into 60-second or 90-second intervals, with aligned subtitles and character-tagged transcripts to facilitate question-answer generation.

\subsubsection{MMSum:}
Multimodal summarization with multimodal output (MSMO) has potential but is hindered by dataset limitations such as poor maintenance and small size. To overcome this, the MMSum dataset \cite{ref76} was developed, featuring human-validated summaries across 17 categories and 170 subcategories, ensuring diverse, high-quality multimodal learning. Benchmarking on MMSum supports video, text, and multimodal summarization tasks.

\subsubsection{LibriSpeech:}
LibriSpeech \cite{ref78} is a 1000-hour read English speech dataset derived from LibriVox audiobooks, sampled at 16 kHz. Designed for training and evaluating speech recognition systems, it includes language-model training data and pre-built models. Models trained on LibriSpeech show superior performance with lower error rates on WSJ test sets compared to those trained directly on WSJ data.

\subsubsection{CATER:}
CATER \cite{ref98} tackles the challenges in video understanding by providing a synthetic dataset designed to evaluate spatiotemporal reasoning with 3D objects and intricate motions. It offers diagnostic tools to assess such models. 

\subsubsection{VoxCeleb:}
VoxCeleb \cite{ref79} is a dataset of over 100,000 audio-visual utterances from 1,251 celebrities, collected via an automated pipeline using YouTube videos, synchronization CNNs for speaker activity, and facial recognition for identity. It includes diverse demographics, professions, and accents, with real-world audio conditions like noise and degradation. Benchmarking shows CNN-based models excel in speaker identification. Unlike SITW, it contains both audio and video, with VoxCeleb2 offering an expanded version.

	\begin{table*}[tbp]
    		\centering
            \caption{Comprehensive summary of the datasets under task specific application with modality, description, size and year of release.}
            \label{tab:tbl_task_specific}
    		\resizebox{\textwidth}{!}{
    		\begin{tabular}{l l c c r}
    			\hline
                \textbf{Dataset} & \textbf{Modality} & \textbf{Description} & \textbf{Size} & \textbf{Time}\\ \hline
                OmniACT & Visuals + Text & A dataset and benchmark for assessing agents’ abilities to generate scripts. & 9802 data points & 2024\\
                InternVid & Video + Text & A large-scale video-centric dataset for video-text representation learning. & 7M+ videos & 2024\\
                SlideVQA & Image + Text & A multi-image document VQA dataset requiring reasoning in slide decks. & 2.6k+ slide decks, 52k+ images & 2024\\
                Peacock dataset & Text + Image & Dataset created for training the Peacock models.& ~916K image-text pairs & 2024\\
                ImageNet & Image & Large-scale object recognition dataset with millions of images. & ~150GB & 2021\\
                CMU MultimodalSD & Text + Image + Audio & Synchronized dataset of video, audio, and text data. & ~12GB & 2023\\
                M3ED & Text + Video & Emotion recognition dataset from videos. & ~35GB & 2022\\
                MELD & Text + Audio & Emotion recognition in dialogues from TV shows. & ~5GB & 2019\\
                TVQA & Text + Video & Question-answering dataset based on TV show videos. & ~200K QA pairs & 2023\\
                CommonVoice & Audio & Multilingual voice recordings for speech recognition. & ~1,000 hours & 2019\\
                AudioCaps & Audio & Audio captioning dataset for sound events. & ~30K audio clips & 2019\\
                HowTo100M & Video + Text & Instructional videos with textual descriptions. & ~100M videos & 2019\\
                EmotionCaps & Text + Audio & Audio clips annotated with emotional descriptions. & ~100K audio clips & 2023\\
                ShapeNet & 3D & 3D models dataset for shape understanding tasks. & ~51K models & 2015\\
                SQA3D & Text + 3D & Spatial question answering in 3D environments. & ~40K scenes & 2022\\
                MMSum & Text & Dataset for multimodal summarization. & ~200K documents & 2023\\
                LibriSpeech & Audio & Corpus of read English speech for ASR. & ~1,000 hours & 2019\\
                VoxCeleb & Audio & Speaker recognition dataset with celebrity voice samples. & ~1,200 hours & 2020\\
                CATER & Video & Action recognition dataset in video content. & ~130K clips & 2020\\

    		\end{tabular}
    	}
    	
    	\end{table*}

\section{Multimodal Datasets for Domain-Specific Applications}

\begin{figure*}[h!]
    	\centering
    	\includegraphics[width=0.8\textwidth]{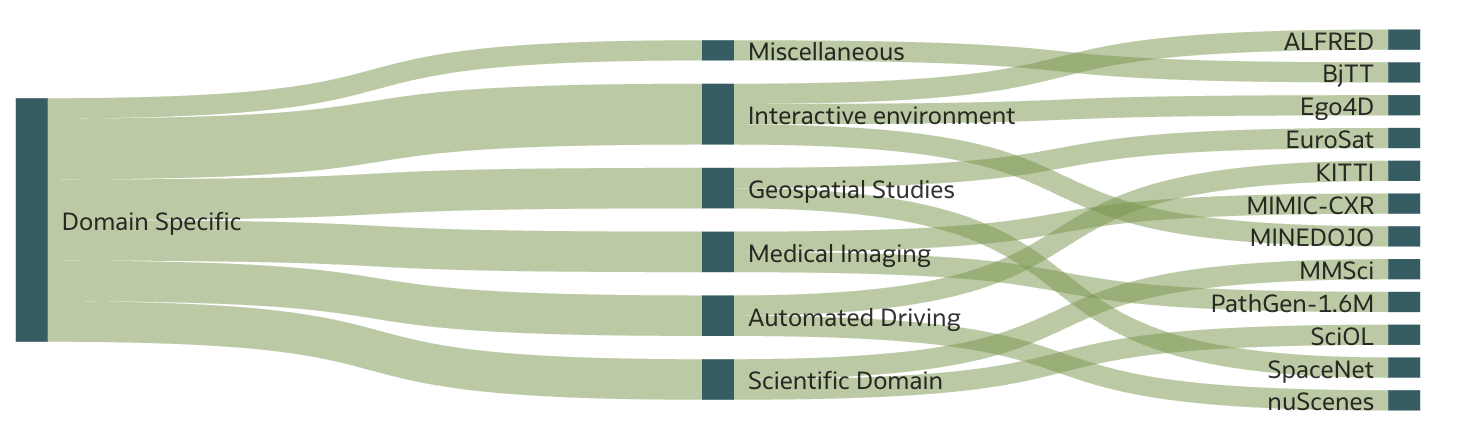} 
    	\caption{Datasets grouped by various domains}
    	\label{fig:domain_split}
        \vspace{-1em}
    \end{figure*}

Recent advancements in large language models have spurred the creation of multimodal datasets tailored to specific domains, complementing existing task-focused datasets. These domain-specific datasets leverage the integration of multiple modalities to address unique challenges across industries. For a detailed overview, refer to Fig. \ref{fig:domain_split} and Table \ref{tab:tbl_domain_specific}.

\subsubsection{Medical Imaging:}
Domain-specific multimodal datasets, particularly in medical imaging, are instrumental for tasks like diagnosis, treatment planning, and patient monitoring. For instance, the MIMIC-CXR dataset \cite{ref99} pairs radiological images with clinical reports, enabling models to understand correlations between visual data and medical language. Similarly, PathGen-1.6M \cite{ref44}, an open-source dataset with 1.6 million image-caption pairs, aids in training models for pathological image analysis, supporting disease detection and prognosis.

\subsubsection{Autonomous Driving:}
Multimodal datasets are essential in autonomous driving, integrating inputs like camera footage, LiDAR, and GPS for robust self-driving systems. Widely used datasets, such as the KITTI Vision Benchmark Suite \cite{ref100} and nuScenes \cite{ref101}, support training models to navigate urban environments, detect objects, and predict road user behavior.

\subsubsection{Geospatial Intelligence:}
Multimodal datasets, integrating satellite images, topographic maps, and textual data, are pivotal in geospatial intelligence for applications like land-use classification, change detection, and disaster response\cite{ref58}. Examples such as SpaceNet \cite{ref102} and EuroSAT \cite{ref103} enable LLMs to derive significant insights by leveraging diverse geospatial information sources.

\subsubsection{Scientific Domain:}
Two prominent datasets enhance scientific figure retrieval tasks. SciOL \cite{ref75}, an extensive open-access corpus, supports multimodal models in scientific research, while MuLMS-Img \cite{ref75} focuses on high-quality image-text pairings in materials science. Both demonstrate significant improvements in figure classification, captioning, and retrieval tasks. Another resource, MMSci \cite{ref77}, compiles data from Nature Communications across 72 disciplines, offering benchmarks for figure captioning and multiple-choice tasks. This dataset reveals substantial model performance gaps, underscoring its potential as a valuable training tool.

\subsubsection{Egocentric and Interactive environment:}

Recent advances in egocentric video datasets have significantly enhanced research in computer vision and natural language processing, particularly for analyzing everyday activities and interactions. Notable contributions include Ego4D \cite{ref86}, featuring 3,670 hours of first-person videos, and ALFRED \cite{ref87}, which links natural language instructions to household action sequences. 

MINEDOJO \cite{ref88} builds on this by offering a framework for developing versatile agents through simulations, a multimodal knowledge base, and flexible agent architectures.

\subsubsection{Miscellaneous:}
The Beijing Text-Traffic (BjTT) dataset \cite{ref91} offers a comprehensive multimodal resource for Intelligent Transportation Systems (ITS) traffic prediction. By integrating textual context with time-series traffic data, it overcomes the constraints of traditional datasets with over 32,000 traffic records paired with descriptive text.

\section{ Dataset Characteristics and Limitations}
The development of advanced multimodal language models relies heavily on large, diverse, and well-annotated datasets. For example, datasets like MS-COCO \cite{ref110} offer extensive real-world data distributions, enabling models to learn intricate cross-modal relationships. Broad modality coverage, as seen in Flickr30k \cite{ref105}, and high-quality annotations enhance training and evaluation effectiveness. Practical applicability is supported by datasets like SpaceNet \cite{ref102}, which align models with real-world tasks. 

\begin{table*}[h!]
    		\centering
            \caption{Comprehensive summary of the datasets under domain specific needs with the dataset type, description, size and year of release.}
             \label{tab:tbl_domain_specific} 
    		\resizebox{\textwidth}{!}{
    		\begin{tabular}{l c c c r }
            \hline
            \textbf{Dataset} & \textbf{Type} & \textbf{Description} & \textbf{Size} & \textbf{Time}\\ \hline
    			
                MIMIC-CXR & Image + Text & Chest X-ray images with radiology reports. & ~370,000 images & 2019\\
                PathGen & Image + Text & A synthetic dataset used for training and evaluating models in generating and understanding complex path trajectories & ~1.6M images & 2023\\
                SciOL & Text & Science-oriented question-answering dataset. & ~1M QA pairs & 2024\\
                MMSci & Text + Image & Multimodal science dataset with images and text. & ~200K instances & 2024\\
                MINEDOJO & Text + Video & Interactive learning dataset from gameplay videos. & ~10K clips & 2022\\
                ALFRED & Text + 3D & Instruction following dataset in 3D environments. & ~100K tasks & 2020\\
                Ego4D & Video & Egocentric video understanding dataset. & ~7K hours & 2022\\
                KITTI & Image + Lidar & Autonomous driving dataset with images and Lidar data. & ~40GB & 2012\\
                nuScenes & Image + Lidar & Self-driving car dataset with urban data. & ~1.4TB & 2020\\
                EuroSat & Image & Satellite image classification dataset. & ~1.5GB & 2019\\
                SpaceNet & Image & Satellite imagery for urban mapping tasks. & ~600GB & 2018\\
                BjTT & Text + Time Series & A dataset for traffic prediction, combining time-series traffic data with descriptive text. & 32,000+ time-series records & 2024

                \end{tabular}
    	}
    		           
    	\end{table*}
However, challenges persist, including data biases, imbalances , and limited diversity in task representation. Moreover, privacy and security concerns underscore the need for responsible dataset use\cite{ref11}. Overcoming these challenges through thoughtful design and curation is vital for advancing robust and ethical multimodal learning.
	
\section{Emerging Trends and Future Dataset Needs}

Multimodal learning has seen significant progress, propelled by advancements in large language models (LLMs) and expansive multimodal datasets. A key trend is the creation of increasingly diverse and complex datasets that mirror real-world scenarios, enhancing the field's potential for addressing practical challenges.

\begin{itemize}
    \item 
    \textbf{Diverse and Geographically Representative Datasets:}
Researchers are advancing multimodal datasets by integrating tactile, olfactory, and physiological signals, while prioritizing geographical and linguistic diversity to enhance generalization and mitigate biases in large-scale models \cite{ref16}.
    \item 
    \textbf{Complex Interactions and Real-World Applications:}
Future multimodal datasets should encompass intricate intermodal interactions while tackling practical applications like healthcare, autonomous systems, and environmental monitoring \cite{ref59}. To ensure transparency, reproducibility, and ethical use, standardized documentation and benchmarking practices are essential \cite{ref63}.
\end{itemize}

\section{Conclusion}
Advancement of multimodal learning relies on development of specialized datasets, which we have grouped into three main categories: training needs, task-specific needs, and domain-specific needs. Training datasets, which are vital for pretraining and instruction tuning methods such as SFT and RLHF, act as the backbone for building and refining multimodal systems.The second category includes needs that are task-specific: these are datasets built for specific tasks that make a model perform well for narrow applications. Finally, domain-specific datasets are essential for solving the peculiarities of specific industries and making the model adaptable and context-sensitive. Although challenges are still there, such as data diversity and effective cross-modal integration, emerging trends in data augmentation and cross-modal learning are mellowing these challenges. Therefore, the development and diversification of these categories of datasets will be highly essential to unlocking the potentials of multimodal systems and further innovations in AI.
%
%
%
 \bibliographystyle{splncs04}
 \bibliography{mybibliography}

\end{document}